\documentclass[journal]{IEEEtran}
\usepackage{comment}

\usepackage{algorithm}

\usepackage{color}

\usepackage{algorithm}
\usepackage{algorithmic}
\usepackage{breqn}
\usepackage{subcaption}
\usepackage{multirow}
\usepackage{psfrag}
\usepackage{url}
\usepackage[colorlinks,citecolor=blue]{hyperref}

\usepackage{cleveref}
\usepackage{booktabs}
\usepackage{rotating}
\usepackage{colortbl}
\usepackage{paralist}
\usepackage{epstopdf}
\usepackage{nag}
\usepackage{microtype}
\usepackage{siunitx}
\usepackage{nicefrac}
\usepackage{fontawesome}
\usepackage{xcolor}
\usepackage{multicol}
\usepackage{wrapfig}
\usepackage{todonotes}
\usepackage{tablefootnote}
\usepackage{threeparttable}

\usepackage{cite}
\usepackage{lipsum}
\usepackage[english]{babel}

\IEEEoverridecommandlockouts

\begin{document}

\title{Improving Federated Aggregation with Deep Unfolding Networks
\thanks{Authors are with School of IT, Deakin University, Australia. email: shiva.pokhrel@deakin.edu.au}
}

\author{Shanika I Nanayakkara, Shiva Raj Pokhrel
and Gang Li
\vspace{-1 cm}
}
\maketitle

\begin{abstract}

The performance of Federated learning (FL) is negatively affected by device differences and statistical characteristics between participating clients. To address this issue, we introduce a deep unfolding network (DUN)-based technique that learns adaptive weights that unbiasedly ameliorate the adverse impacts of heterogeneity. The proposed method demonstrates impressive accuracy and quality-aware aggregation. Furthermore, it evaluated the best-weighted normalization approach to define less computational power on the aggregation method. The numerical experiments in this study demonstrate the effectiveness of this approach and provide insights into the interpretability of the unbiased weights learned.
 By incorporating unbiased weights into the model, the proposed approach effectively addresses quality-aware aggregation under the heterogeneity of the participating clients and the FL environment. Codes and details are \href{https://github.com/shanikairoshi/Improved_DUN_basedFL_Aggregation}{here}.
\end{abstract}

\newtheorem{definition}{Definition}[section]
\newtheorem{example}{Example}[section]
\newtheorem{theorem}{Theorem}[section]
\newtheorem{proposition}{Proposition}[section]
\newtheorem{corollary}{Corollary}[section]
\newtheorem{lemma}{Lemma}[section]

\section{Introduction}\label{sec-intro}

Federated learning (FL), initially introduced by Google in \cite{mcmahan2017communicationFedAvg}, revolutionizes collaborative training of machine learning models using data from numerous participating devices, called clients, while ensuring that the privacy of local data remains intact. Google's FL consists of a central server, which iteratively incorporates each client's local update and aggregates the updates from all clients \cite{wang2021field}.

The fundamental assumption of voluntary collaboration and data sharing by all clients in FL (initially discussed in~\cite{pokhrel2020federated, pokhrel2020federated1}) throughout the training process may not always hold, particularly for clients with isolated data who prioritize their own interests and require incentives or rewards for participation.
How to handle the impact of heterogeneity, intermittent client availability, and sparcity, under aggregation of FL remains partially solved.
For example, new weighting aggregation rules~\cite{zhao2022dynamic} have emerged as the simplest approach to solving these limitations.
Consequently, numerous studies have been dedicated to devising effective approaches to determine the appropriate weights by employing techniques such as clustering~\cite{sattler2020clustered}, Shapley~\cite{tang2021FedSV}, graph learning~\cite{xing2022big} and reweighting~\cite{zhao2022dynamic} including others.
However, all the aforementioned works~\cite{zhao2022dynamic, sattler2020clustered, tang2021FedSV, xing2022big} lack scalability, suffer from poor unbiasedness, and involve a high computational cost.

\textcolor{black}{
In the field of signal processing and wireless communications, the integration of deep unfolding network (DUN) for exploiting model-based deep learning, has been quite successful for various applications~\cite{monga2021algorithm, 
sabulal2020joint}. We find ways to extend DUN as a tool to adjust the weights in FL for the desired level of optimized aggregation. To this end, by employing DUN, we will study ways to assign weights that can be optimized dynamically, allowing for adaptive and heterogeneous weight assignments. This learning-to-learn approach shows promise in improving the performance of FL by surpassing heuristic weighting methods. Numerical experiments demonstrate that deep unfolding can effectively adapt the weights to heterogeneous scenarios, resulting in improved test accuracy. } 

{DUN offers advantages over traditional heuristic methods, and we employ DUN to explore and exploit the rooms for refinement in FL aggregation by achieving optimal weight assignments. By diving deeper into the intricacies of weight tuning, it is possible to enhance the performance of the FL with DUN. Such an approach with DUN will involve exploring more advanced descent techniques or algorithms to precisely adjust the weights, with the aim of dynamically maximizing the overall effectiveness of the FL process. Therefore, while the adoption of DUN in weight optimization is a significant step forward, there is an ongoing non-trivial challenge that involves a major rethink and original development to unlock the full potential of DUN for precise weight tuning tightly coupled with FL dynamism.}

 Our primary focus is on the weighting strategy, using relevant findings from~\cite{nakai2022Duw}. The aim is
to overcome the major limitations of existing work, including aspects such as lack of scalability, poor unbiasedness, and computational cost.
Our experimental results demonstrate the effectiveness, unbiasedness, and robustness of our proposed approach.

Our main contributions in this paper are
\begin{itemize}
    \item\textcolor{black}{ We extend DUN to precisely optimize the
weight assignments in a more fine-grained manner, aiming to maximize the overall effectiveness of the
FL process tackling computational cost.}
    \item \textcolor{black}{We tackle heterogeneity by enhancing the unbiasedness in heuristic DUN weighting strategies.}
    \item \textcolor{black}{Through experimental validation, we demonstrate the effectiveness and potential of precise parameter tuning in FL using DUN and unlock new aggregation dimensions.}
\end{itemize}

The remainder of this paper is structured as follows.
 Section \ref{sec-gaps} provides existing approaches and significant gaps.
Section \ref{sec-preliminaries} provides a general description of FL and deep unfolding in FL,
Section \ref{sec-problem} provides problem formulation around the given research area that causes this paper.
Section \ref{sec-experiment} discusses the goals, methodologies, and results of the proposed research design with analysis, and in addition, Section \ref{sec:conclusion} provides the conclusions and future directions of this research.

\section{Literature and Preliminaries} \label{sec-preliminaries}

\subsection{Related Work}
\label{sec-gaps}

Federated Averaging (FedAvg)~\cite{mcmahan2017communicationFedAvg} is the pioneering standard in federated aggregation in which weights are typically determined based on the number of local data samples available to each client.
However, the size of a client's data alone does not guarantee a significant contribution to performance. In particular, when certain clients possess small but distinct datasets that are not present in other clients, it is important to prioritize their information in the aggregation. This emphasis is crucial for developing a highly accurate model that can capture unforeseen distributions~\cite{nakai2022Duw}. On the other hand, clients with large datasets but limited computational capabilities and/or unreliable conditions should never be given a higher weight. Therefore, it is essential to consider the heterogeneity of the clients when determining the aggregation weights.

Some of the aggregation approaches focused on contribution evaluations. 
For example, the Federated Shapley Value (FedSV)~\cite{tang2021FedSV}, dynamically
updates the global model aggregation weights according to each
client contribution in each round;
Guided Truncation Gradient Shapley (GTG Shapley)~\cite{liu2022gtg} presents a contribution evaluation strategy in horizontal FL settings by replacing submodel retraining with reconstruction; Contribution-based device selection modifies the weighting strategy using a modified truncated Monte
Carlo~\cite{pandey2022contribution}. 

Others determine the robustness and fairness to define the corresponding weights~\cite{cao2020fltrustcosinesimilarity, jhunjhunwala2023fedexp, malinovsky2022serverstepsizenastya}.

Cao \emph{et al.}~\cite{cao2020fltrustcosinesimilarity} presented FLTrust, which evaluates the cosine similarity of angles between two vectors: local updates, and server updates obtained from a separate global training data set, and defines trust scores to identify unreliabilities through weighting. FedExp~\cite{jhunjhunwala2023fedexp} introduces a novel aggregation rule in FL that dynamically determines the size of the server step based on an extrapolation mechanism inspired by Projection Onto Convex Sets (POCS). They employ ideas from~\cite{malinovsky2022serverstepsizenastya} and emphasize the importance of assigning smaller server steps in highly heterogeneous FL client systems. FedExp analyzes and adjusts the server step size in an adaptive manner, considering individual client progress and heterogeneity levels.

The recently proposed dynamic reweighting technique (DR) \cite{zhao2022dynamic} aims to promote fairness in FL by assigning higher weights to clients with higher losses. Fairness is defined in terms of achieving a more uniform distribution of accuracy among model parameters, indicating a fairer solution for the desired objective. 

DUN has been successfully implemented in dynamic parameter tuning under stringent iteration constraints, which can be utilized within FL to encourage learning of hyperparameters such as client weights, client and server step sizes, and many others. For instance, Kasai and Wadayama~\cite{nakai2022Duw} propose a DUN-based weighting strategy for FL.

\subsection{Federated Learning and Deep Unfolding Networks}

We adopt the following generalized objective function to capture the dynamics of the data flow and local updates. 
If the number of clients is indicated by $K$, the training data sets as {$D\textsubscript{1}.........D\textsubscript{k}$}, training samples as $D\textsubscript{1}={(x\textsubscript{1},y\textsubscript{1}),....(x\textsubscript{n},y\textsubscript{n})} $
iterations as $ t \in {{0, 1, ...., R-1}}$, the server model as $w$ and the client model as $w \textsubscript{k}$, the local objective can be shown as

\begin{equation}
        f(w \textsubscript{k}) = \frac{1}{n} \eta\textsubscript{k} \sum\limits_{x,y\in D\textsubscript{k} }L(x,y ;w)
\label{eq:generalLocal}
\end{equation}
For the update of the local model of $\Delta w\textsubscript{k}\textsuperscript{t} = f(w\textsubscript{k} \textsuperscript{t}) -  f(w \textsuperscript{t})$, the global aggregation can be shown as a non-negative scalar $\theta\textsubscript{k}=\sum_{j=1}^J \frac{|D\textsubscript{j}|}{\sum_{j\in N}|D\textsubscript{j}|}, \theta\textsubscript{k} \in \Re$, representing the weight of the client k.

\begin{equation}
       w\textsuperscript{t+1} \longleftarrow 
    w\textsuperscript{t} + \eta\textsubscript{g}
   \theta\textsubscript{k}
    \Delta w\textsubscript{k}\textsuperscript{t}
\label{eq:generalGlobal}
\end{equation}

DUNs involve exploiting and reformulating an iterative algorithm as a deep neural network~\cite{gregor2010learning} to develop a solution, which is gradually refined through a series of iterations, each building on the previous one. DUN exploits this iterative process and represents it as a neural network with multiple layers. In DUN, each layer of the neural network corresponds to an iteration of the original algorithm. The input to each layer is the output of the previous layer, and the network parameters are learned by training to mimic the behavior of the iterative algorithm~\cite{balatsoukas2019deep}. 

Consider an iterative algorithm for solving a problem, which can be denoted as:
$x^{(k+1)} = F(x^{(k)}, \theta^{(k)}) $     
In this notation:
$x^{(k)}$ represents the state or solution in iteration k.
$F(x^{(k)},\theta^{(k)})$ denotes the function or operation applied to the state $x^{(k)}$ using parameters $\theta^{(k)}$ to obtain the updated state $x^{(k+1)}$.
Deep unfolding can transform such an iterative algorithm into a deep neural network as
$
x^{(k+1)} = NN(x^{(k)}, \theta^{(k)})
$, where $NN(x^{(k)}, \theta^{(k)}) $ represents a deep neural network that takes the state $x^{(k)} $ and the parameters $\theta^{(k)} $ as input and generates $x^{(k+1)}$.

\section{Proposed DUN for FL Aggregation}

\begin{figure}[b]
  \centering
  \includegraphics[width=0.459\textwidth]{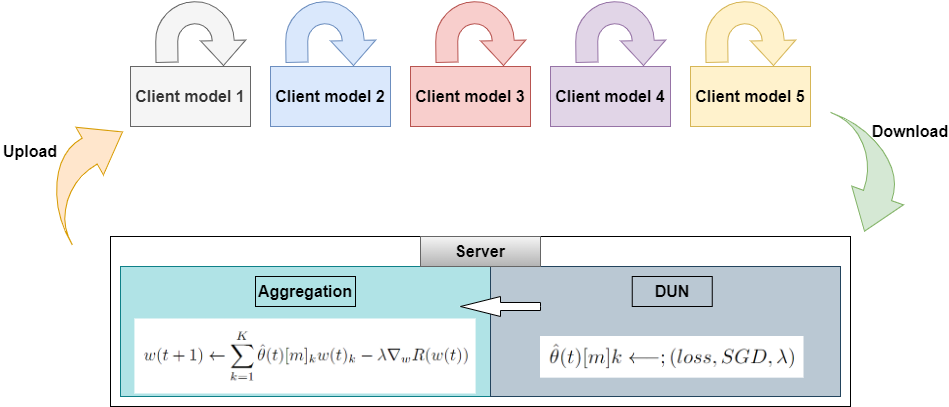}
  \caption{A High-level view of the proposed approach.}
  \label{fig:blockdigram}
\end{figure}

We explore and exploit DUNs under FL to address the challenges associated with heterogeneity and improve the performance of the learning process. An abstract view of the proposed approach is shown in Fig.~\ref{fig:blockdigram}. By incorporating deep unfolding techniques into FL aggregation, we develop adaptive weighting that can be learned to account for variations in devices and statistical characteristics among the participating clients. Our proposed DUN of FL involves iteratively updating the local models at each client and aggregating the updates at a central server. The DUN process unfolds the iterative optimization procedure and incorporates it into the learning framework (as explained in~\cite{mogilipalepu2021federated}). 

By plugging DUN, we enable joint optimization of the model parameters and adaptive weights during aggregation that captures the qualitative heterogeneity of the clients. The new theory is important because the models are effectively adapted to the diverse (quality-driven) characteristics of the participating clients, leading to improved performance and accuracy. The learned weights can account for differences in data distributions, device capabilities, and statistical properties, ensuring a more robust and fair FL process. Furthermore, the results of the DUNs are fed to improve the convergence of the FL process, mitigate the impact of heterogeneity, and improve the performance and precision of the model in this collaborative training setting, along the lines of~\cite{hadou2023stochastic}.

The proposed DUN in aggregation, shown in Fig.~\ref{fig:blockdigram}, constantly seeks to address several impeding problems in FL including the following main issues.

\noindent 1. 
Allows for the precise optimization of weight assignments in a more fine-grained manner. This enables better utilization of client contributions and improves the overall effectiveness of the FL process.

\noindent 2. Addresses heterogeneity challenges by improving unbiasedness in heuristic weighting strategies. By considering the varying characteristics of clients, DUN ensures a fair and balanced representation of client data during the learning process. 

\noindent 3. Minimizes computational costs in FL by optimizing weight assignments with greater precision, thus reducing unnecessary computational burdens on resource-constrained clients. Overall, DUN provides a powerful framework that contributes to solving key problems in FL, improving its performance and scalability.

 By integrating the identification of clients that are not available in the DUN model and incorporating it as part of the FL optimization process, we leverage the power of deep learning to effectively detect and identify clients that are currently unavailable for participation.

\subsection{Problem Formulation}
\label{sec-problem}

The aggregation step in FL requires a well-designed weighting strategy, which remains an active area of research. Existing approaches have limitations in effectively handling participant heterogeneity and client availability. Recently, a deep unfolding-based weighting strategy has shown promise in improving weight calculation compared to other methods. However, it still has significant drawbacks such as high computational costs for preprocessing, limited communication improvement, instability on unreliable clients, and scalability issues.

In this paper, we present a comprehensive study aimed at improving precision while maintaining computational and communication efficiency in the weighting process. Our experiments focus primarily on investigating optimization and regularization approaches to tackle these challenges effectively.

Furthermore, we aim to propose an optimal and unbiased weighting mechanism using deep unfolding. Additionally, we demonstrate how computational time can be reduced by implementing appropriate techniques within the deep unfolding process.

\begin{table*}
 \begin{center}
\caption{Experimental settings Details and Evaluation}
\begin{tabular}{ *{5}{c} }
\hline

\textbf{Settings}&  \textbf{Heterogeneity}& 
\textbf{Naive Iterations}&\textbf{Our Iterations}&\textbf{Complexity Reduction}\\
\hline
\hline
Setting 01&Statistical&400 &100&30\%\\

Setting 03&Computation &500&100&40\%\\

Setting 04&Communication &800&100&70\%\\
\hline
\bottomrule
\end{tabular}
\label{table:efficiency}
 \end{center}
\end{table*}

\subsection{Algorithm Design}

\begin{algorithm}[b]
\caption{DeepUnfolding with Regularization}
\begin{algorithmic}[1]
\REQUIRE $M, T, K, \lambda$
\STATE Initialize model parameter $w(0)$
\STATE \textbf{for} $m = 0$ : $M$,\; $[m] \leftarrow 0$
    \STATE \quad \quad \textbf{for} each client $k$
      \\ \quad \quad \quad $w(t)_k \leftarrow \text{Client-Update}(k, w(t))$ \quad (\textit{loss calculation})
      \STATE \quad \textbf{for} $t = 0$ : $T$\\ 
    \quad \quad Distribute $w(t)$ to all clients
  \\ \quad \quad Server-Update parameters\\\;\;\;\quad $w(t+1) \leftarrow \sum_{k=1}^{K} \hat{\theta}(t)[m]_k w(t)_k - \lambda \nabla_w R(w(t))$\\
 \quad\quad { $[m] \leftarrow [m] + \sum_{t=0}^{T-1} \sum_{k=1}^{K} t_k$}
  \\ \quad \quad Server-Update $\hat{\theta}(t)[m]_k \rightarrow \hat{\theta}(t)[m+1]_k$\\\;\;\;\; \quad (SGD with $[m]$, weight decay $\lambda$)
\RETURN $\{\hat{\theta}(\textsuperscript{t})[M]_k\}_{T-1, K, t=0, k=1}$
\end{algorithmic}
\label{algo:shanika}
\end{algorithm}

We develop an improved DUN-based FL aggregation rule by efficiently increasing the unbiased client contribution.
For that, we implement appropriate optimization with penalty regularization over the deep unfolding training layers. The details of our design are outlined in Algorithm~\ref{algo:shanika}.

\textcolor{black}{
The learnable parameters, as shown in Algorithm~\ref{algo:shanika}, are the weights $\hat{\theta}\textsuperscript{t}\textsubscript{k} $ applied with regularization in each round $t$. Intuitively, we modify the weighted averaging procedure at the server in \eqref{eq:generalGlobal} as }
\begin{equation}
       w\textsuperscript{t+1} \longleftarrow 
    w\textsuperscript{t} + \eta\textsubscript{g}
   \hat{\theta}\textsuperscript{t}\textsubscript{k}
    \Delta w\textsubscript{k}\textsuperscript{t} :\hat{\theta}\textsuperscript{t}\textsubscript{k}\longleftarrow{(SGD,\lambda)}
\label{eq:ProposedEq}
\end{equation}
In the context of FL, obtaining relevant training data that align with the specific situation is crucial to learning the weights. Clients can provide some or all of their data to train the consequences when cooperating. It is essential to highlight that the data can remain on each client, and privacy is preserved by performing backpropagation over the air. Although this concept can be applied to different FL algorithms and incorporate various operations, this paper primarily focuses on presenting a comprehensive learning procedure tailored to a specific scenario. In this scenario, the system involves a relatively small number of devices that remain in operation for an extended period.

In $m$ -th, $T$ FedAvg rounds are executed with the weights ${\hat{\theta}(t[m] k } (t = 0,. . , T-1)$. The losses accumulated during the process are sent to the server. This weighting mechanism is summarized in Algorithm~\ref{algo:shanika}. The weights are updated after T rounds using a standard Stochastic Gradient Descent (SGD), including weight decay. Our proposed approach incorporates the learned weights based on deep unfolding with regularization coefficients ${\hat{\theta}(t)[m] k;\lambda }$.

As shown in Algorithm~\ref{algo:shanika}, step 4, the weight decay term is updated to $\lambda \nabla_w R(w(t))$, where $\lambda$ represents the weight decay coefficient and $\nabla_w R(w(t)) $represents the gradient of the weight decay regularizer function $R(w(t))$ regarding the model parameters $w(t)$.

The proposed weight decay is a regularization method that can enhance the ability to carry out training in FL to generalize well. It addresses the problem of overfitting by encouraging the model to focus on basic patterns and trends that are common between clients rather than memorizing noise or client-specific details. This is achieved by incorporating a penalty term into the loss function, which promotes smaller weights and reduces the complexity of the model. Weight decay is particularly effective when the unfolding network has many parameters.

\subsection{Originality and novelties}
Unfolding networks in FL often have a high capacity and may be prone to overfitting local client data. Weight decay helps alleviate this issue by discouraging the model from assigning excessive weights to specific features or patterns that may be unique to individual clients. Indeed, it encourages a more balanced weight distribution, enabling the model to generalize well to new, unseen data.

The proposed weight-decay regularization is novel in enhancing the generalization performance with DUN by mitigating overfitting. It enables the model to capture essential shared patterns among clients, improving performance when evaluating the FL model aggregation over unseen consequences. Such a regularization technique is particularly advantageous during the aggregation of client weights in FL. Our experimental evidence explained later suggests that weight decay can achieve higher accuracy while maintaining high computation efficiency in most FL environments.

Our adoption of the SGD optimizer instead of adaptive optimization allows original benefits.
Specifically, it enables better control over the training process.
In DUNs, the unfolding process involves iterative optimization steps. Using SGD allows for more fine-grained control over the learning process since we can explicitly specify the learning rate and adjust it as needed. Such control can achieve desirable convergence behavior and balance the trade-off between convergence speed and FL performance.

Adaptive learning rate optimization is another novel aspect. We develop a mechanism to adjust the learning rates for individual parameters based on their historical gradients. While this adaptivity can be advantageous in many cases, it may introduce instability or undesirable behavior in the DUN process. We find novel ways which involve unfolding a recurrent or iterative process and maintaining a stable learning rate throughout the unfolding iterations, which are essential for improving the proper convergence rate.

Our adaptive learning rate design maintains additional statistics and momentum terms for each parameter, resulting in higher memory requirements and computational overhead than SGD. This additional computational burden can become significant in DUNs, where the unfolding process may involve numerous iterations. This is balanced in our design with SGD, which has a more straightforward update rule and requires a few memory resources, making it more efficient in such scenarios. Indeed, DUN tasks will soon become complex, and using a simpler optimization algorithm like SGD often helps to reduce the complexity and make it easier to understand, analyze, and scale the dynamics of the underlying unfolding process.

\section{Experiment and Analysis} \label{sec-experiment}

We evaluate the performance using numerical experiments.
The number of clients and rounds was set at $K = 5, M = 100$ and $T = 10$, respectively. We develop three experimental settings as given in Table \ref{table:efficiency} and use three characteristic training datasets extracted from the MNIST dataset. For tractability and benchmarking, continuing the lines of~\cite{nakai2022Duw}, we have the following settings: i) \textit{Setting 01} contains imbalanced data that introduce statistical heterogeneity; ii) \textit{Setting 02} skews computational heterogeneity by adjusting a number of epochs while maintaining balanced data; and iii) \textit{Setting 03} generates communication heterogeneity into data.

Our objective is to observe and study the effectiveness and appropriateness of the solution to improve the quality-aware efficiency of federated aggregation, where we employ deep unfolding with a regularization penalty. We evaluate in the three environments, where we maintain different settings for various heterogeneity aspects. 

\begin{figure}[h]
  \centering
  \begin{subfigure}[b]{0.2132\textwidth}
    \includegraphics[width=\textwidth]{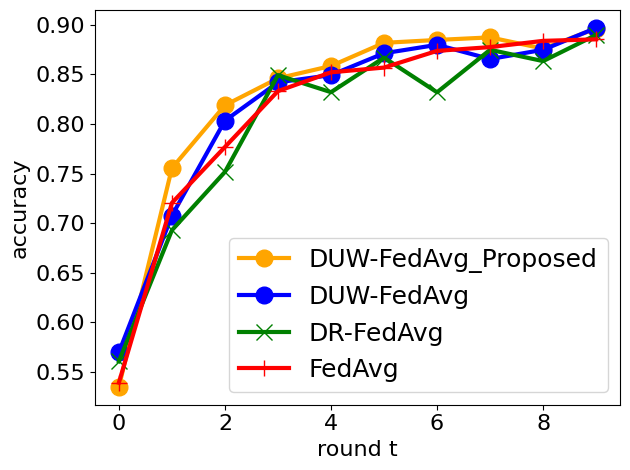}
    \caption{Setting 01: Accuracy}
    \label{fig:set1Acc}
  \end{subfigure}
  \hfill
  \begin{subfigure}[b]{0.2132\textwidth}
    \includegraphics[width=\textwidth]{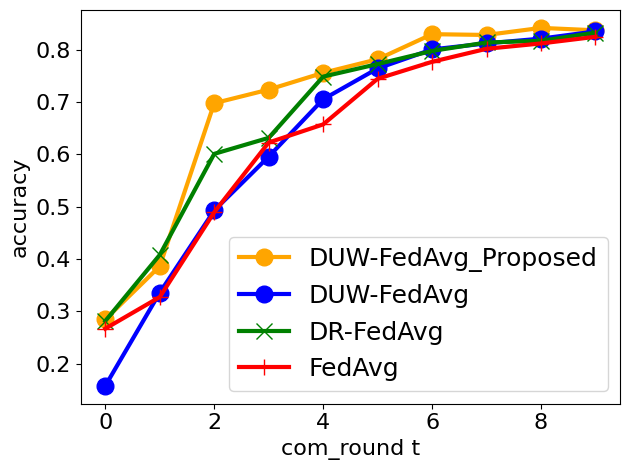}
    \caption{Setting 02: Accuracy}
    \label{fig:set2Acc}
  \end{subfigure}
  \hfill
  \begin{subfigure}[b]{0.2132\textwidth}
    \includegraphics[width=\textwidth]{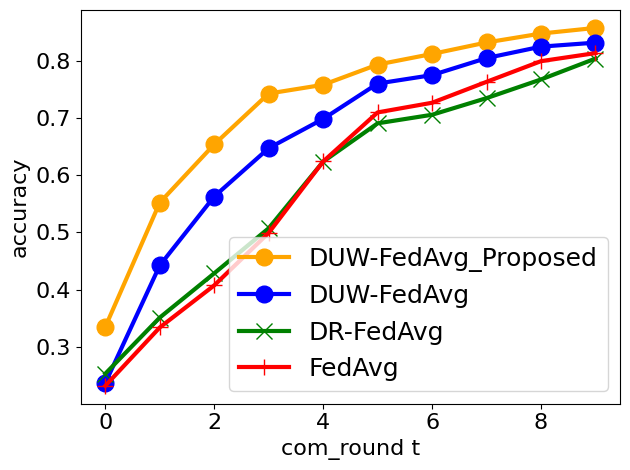}
    \caption{Setting 03: Accuracy}
    \label{fig:set3Acc}
  \end{subfigure}
    \hfill
  \begin{subfigure}[b]{0.2132\textwidth}
    \includegraphics[width=\textwidth]{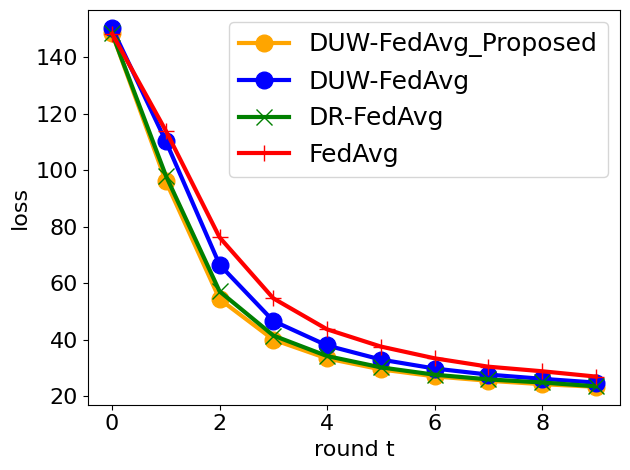}
    \caption{Setting 01: Loss}
    \label{fig:set1Loss}
  \end{subfigure}
  \hfill
  \begin{subfigure}[b]{0.2132\textwidth}
    \includegraphics[width=\textwidth]{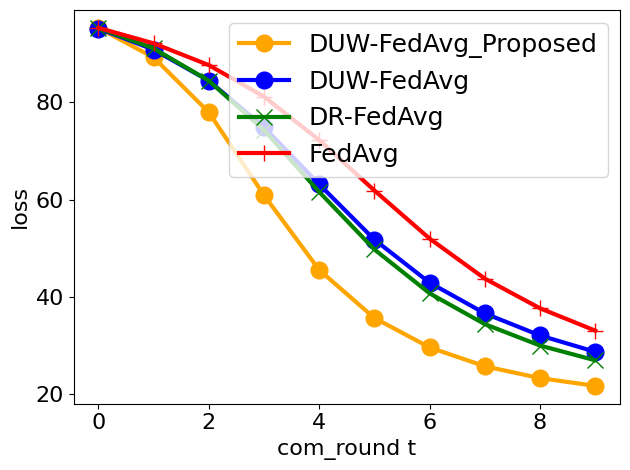}
    \caption{Setting 02: Loss}
    \label{fig:set2Loss}
  \end{subfigure}
  \hfill
  \begin{subfigure}[b]{0.2132\textwidth}
    \includegraphics[width=\textwidth]{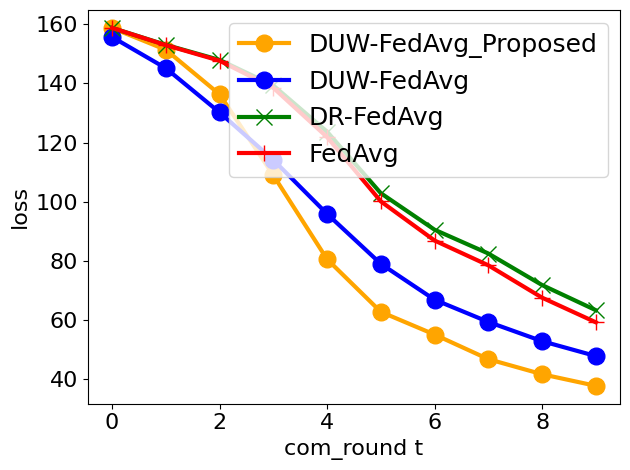}
    \caption{Setting 03: Loss}
    \label{fig:set3Loss}
  \end{subfigure}

  \caption{Accuracy and Losses for various heterogeneous setups, (ia): Communication, (iib): Computation, (iiic): Statistical}
  \label{fig:Main_AccLoss}
\end{figure}

\begin{figure}
  \centering
  \begin{subfigure}[b]{0.2132\textwidth}
    \includegraphics[width=\textwidth]{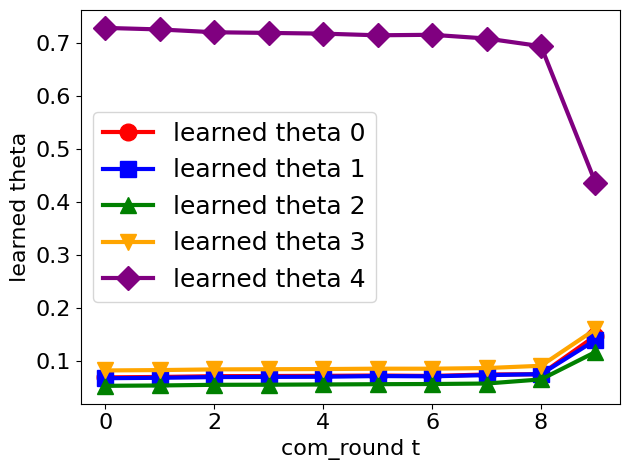}
    \caption{Setting 01}
    \label{fig:set1Org}
  \end{subfigure}
  \hfill
  \begin{subfigure}[b]{0.2132\textwidth}
    \includegraphics[width=\textwidth]{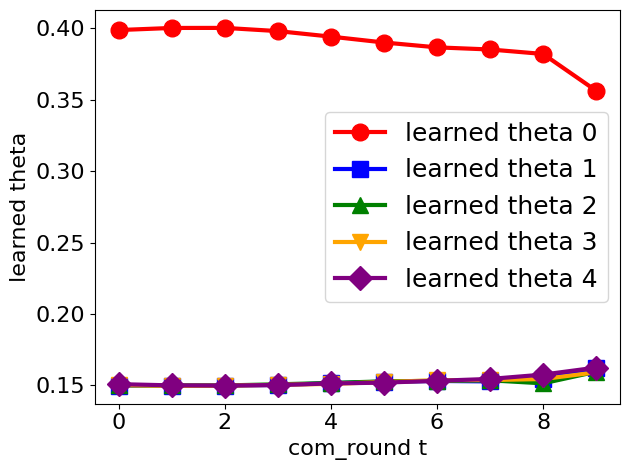}
    \caption{Setting 02}
    \label{fig:set2Org}
  \end{subfigure}
  \hfill
  \begin{subfigure}[b]{0.2132\textwidth}
    \includegraphics[width=\textwidth]{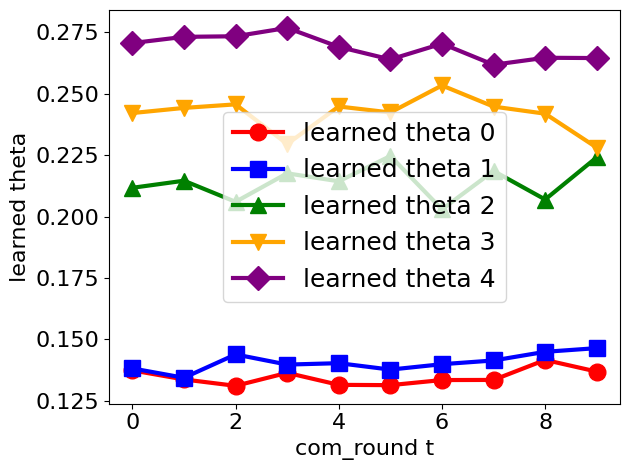}
    \caption{Setting 03}
    \label{fig:set3Org}
  \end{subfigure}
    \hfill
  \begin{subfigure}[b]{0.2132\textwidth}
    \includegraphics[width=\textwidth]{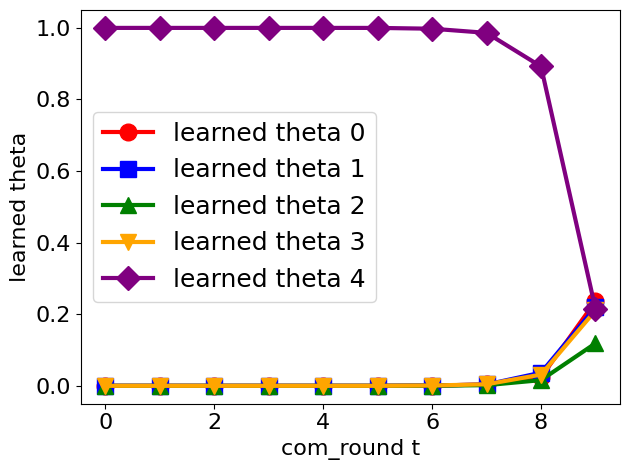}
    \caption{Setting 01}
    \label{fig:set1Our}
  \end{subfigure}
    \hfill
  \begin{subfigure}[b]{0.2132\textwidth}
    \includegraphics[width=\textwidth]{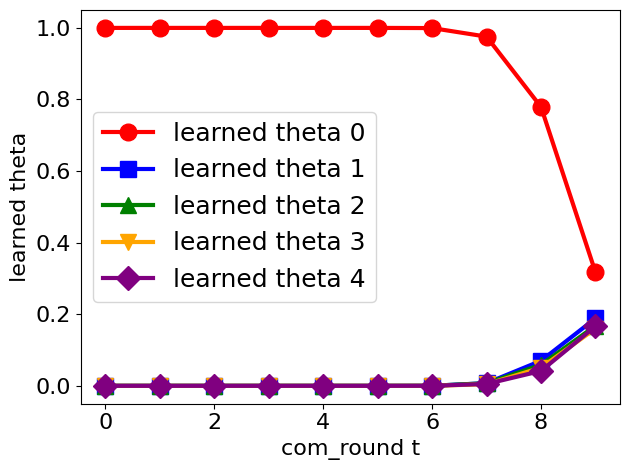}
    \caption{Setting 02}
    \label{fig:set2Our}
  \end{subfigure}
  \hfill
  \begin{subfigure}[b]{0.2132\textwidth}
    \includegraphics[width=\textwidth]{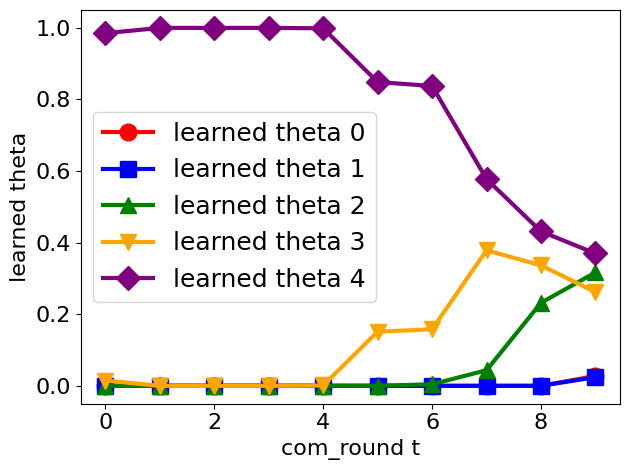}
    \caption{Setting 03}
    \label{fig:set3Our}
  \end{subfigure}
  \caption{Comparison between Learned Weights of original work and proposed work.} 
  \label{fig:main_LearnedWeights}
\end{figure}

Figures \ref{fig:Main_AccLoss} a)-c) show accuracy and Figures \ref{fig:Main_AccLoss} d)-f) show loss behaviour over the communication rounds for selected experimental setups given in Table \ref{table:efficiency}.
significantly,
We trained our deep unfolding network only for 100 layers/ iterations, and our results show that it has achieved high accuracy with the lowest loss within limited iterations for the weight learning process.
The result shows that the original work has less accuracy and high loss if they learned their weights only for 100 layers/ iterations. \footnote{In a similar context, \cite{nakai2022Duw} uses 400-800 layers/ iterations for their experiments to show their accuracy over the other methods, such as DR_FedAvg, and FedAvg.}

 For benchmarking, we generate Figures~\ref{fig:main_LearnedWeights} a)-c) using the proposed approach from \cite{nakai2022Duw}. They show that the distribution of model parameters among clients is not fair in the three settings of Table \ref{table:efficiency}. With the proposed approach, our experimental results, depicted in Figures~\ref{fig:main_LearnedWeights} d)-f) demonstrate the unbiased behavior of our method in various heterogeneous setups. Observe that in the final round, the model parameters of all clients in our proposed method are qualitatively and fairly aggregated. 

\section{Conclusion} \label{sec:conclusion}

We have proposed an improved and computationally efficient deep unfolding network (DUN)-based weighted averaging for FL aggregation. Our study involved evaluating the performance with the proposed weights in specific environments and with varying numbers of reliable clients. The results demonstrated that our proposed approach achieved higher accuracy considering computational constraints compared to traditional methods. The learned weights showed adaptability to capture qualitative insights from the heterogeneity present in the federated network system. With DUN, we have successfully developed an important understanding that a realistic FL environment often involves many unreliable and sporadic clients, which requires extensive future research challenges.

\bibliography{report-ieee.bib}
\bibliographystyle{IEEEtran}

\end{document}